\pdfoutput=1

\documentclass[11pt]{article}

\usepackage[]{emnlp2021}

\usepackage{times}
\usepackage{latexsym}
\usepackage{titlesec}
\titleformat{\subsubsection}[runin]
{\normalfont\bfseries}{\thesubsection}{1em}{}
\usepackage{multicol}
\usepackage{multirow}
\usepackage{booktabs}
\usepackage{amsmath}
\usepackage{adjustbox}
\usepackage{comment}
\usepackage{caption}
\usepackage{subcaption}

\usepackage[T1]{fontenc}

\usepackage[utf8]{inputenc}

\usepackage{microtype}

\title{Does Vision-and-Language Pretraining Improve Lexical Grounding?}

\author{
    Tian Yun \and Chen Sun \and Ellie Pavlick \\
    Brown University \\
    \texttt{\{tian\_yun, chensun, ellie\_pavlick\}@brown.edu} \\
}

\begin{document}
\maketitle

\begin{abstract}
Linguistic representations derived from text alone have been criticized for their lack of \textit{grounding}, i.e., connecting words to their meanings in the physical world. Vision-and-Language (VL) models, trained jointly on text and image or video data, have been offered as a response to such criticisms. However, while VL pretraining has shown success on multimodal tasks such as visual question answering, it is not yet known how the internal linguistic representations themselves compare to their text-only counterparts. This paper compares the semantic representations learned via VL vs.\ text-only pretraining for two recent VL models using a suite of analyses (clustering, probing, and performance on a commonsense question answering task) in a language-only setting. We find that the multimodal models fail to significantly outperform the text-only variants, suggesting that future work is required if multimodal pretraining is to be pursued as a means of improving NLP in general. 

\end{abstract}
\section{Introduction}
Large pretrained language models (LMs)--e.g., BERT \citep{devlin-etal-2019-bert}, GPT \citep{radford2019language,DBLP:journals/corr/abs-2005-14165}-- derive representations of words and sentences by distilling patterns that exist in large text corpora. While such representations have shown strong empirical performance on many benchmark language understanding tasks, they have been criticized for their lack of \textit{grounding}, i.e., the ability to connect words to the real-world entities, events, and ideas to which they refer. While grounding is obviously necessary for mulitmodal language understanding tasks (e.g., identifying a dog in an image), it has further been argued to be fundamental for learning semantic representations in general. For example, \citet{bender-koller-2020-climbing} argues that models trained without grounding will ultimately fail on some text-only tasks such as goal-oriented dialogue, and \citet{DBLP:journals/corr/abs-2104-10809} argues that an embedding space learned from text alone cannot encode the correct conceptual structure.
One proposed solution is to shift from text-only models to multimodal models, which learn to associate language with representations of the non-linguistic world \citep{bisk-etal-2020-experience}. Such approaches are intuitively appealing, but have not yet been rigorously analyzed in practice.

We test the hypothesis that grounded pretraining yields better linguistic representations (of words and sentences) than does text-only pretraining. For two recently released vision-and-language (VL) models, VideoBERT and VisualBERT, we compare the performance of the multimodal model to a text-only variant. We measure how well the representations encode 1) common sense inferences about the physical world, 2) the semantic structure of verbs and their arguments, and 3) compositional information about objects and their properties.  Overall, we do not find evidence that the linguistic representations learned via multimodal pretraining differ meaningfully from those learned from text alone. We argue that such results do not imply that grounding is unimportant for language understanding, but rather that substantial future work on how to combine modalities is required if multimodal methods are to impact NLP in general. Our code is available at \url{https://github.com/tttyuntian/vlm_lexical_grounding}.

\section{Related Work}

\paragraph{Analyzing Pretrained LMs.} 
There has been substantial prior work on analyzing pretrained LMs and the linguistic properties of their representations, looking, e.g., at syntactic parse structure \citep{hewitt-manning-2019-structural,linzen2016assessing}, semantic structure such as semantic roles and coreference \citep{tenney-etal-2019-bert}, lexical semantics \citep{chronis-erk-2020-bishop,vulic-etal-2020-probing}, and lexical composition \citep{yu-ettinger-2020-assessing}. Particularly relevant to our studies is prior work which has explored how well text-only models capture commonsense knowledge about the physical world via intrinsic \citep{ettinger2020,DBLP:journals/corr/abs-1908-02899} and extrinsic \citep{zellers-etal-2018-swag,zellers-etal-2019-hellaswag,bisk2020piqa} measures. Despite the interest in representations of the non-linguistic world, such analyses have not, to our knowledge, been run on multimodal LMs.

\paragraph{Vision-and-Language Pretraining.} There is a long history of multimodal distributional semantics models \citep{howell2005model,lazaridou-etal-2015-combining}, to which pretrained transformer-based models are the latest addition \citep{sun2019videobert,li-etal-2020-bert-vision}. Evaluations of these recent vision-and-language (VL) models has tended to focus on inherently multimodal tasks
, e.g., image and video captioning \citep{sun2019videobert}, visual question answering \citep{li-etal-2020-bert-vision}, or instruction following in robotics \citep{vln_bert}.
\citet{cao2020behind} describes a series of ``probing'' analyses for multimodal language representations, but focuses on explicit grounding, e.g., to where do models attend in the image when processing ``dog''? Little work has analyzed whether the presence of grounded training data impacts the linguistic representations in general. Work that does perform exploratory analyses of the multimodal conceptual representations \citep{tan-bansal-2020-vokenization,Radford2021LearningTV} does not include analysis of comparable text-only models, limiting the conclusions that can be drawn. 

\section{Vision-and-Language Pretraining}

This section describes pretraining approaches that use both vision and language information. In particular, we focus on two that extend the BERT~\citep{devlin-etal-2019-bert} pretraining for text, VideoBERT~\citep{sun2019videobert} and VisualBERT~\citep{li-etal-2020-bert-vision}. Both are single-stream models which directly combine visual and text information at the model inputs, and are trained on paired video+speech and image+caption data, respectively.

More specifically, VideoBERT encodes video data by vector quantization, mapping visual features extracted from 1.5 seconds long video segments into ``visual words'' with K-Means clustering. The authors downloaded around 300K publicly available cooking videos from YouTube, and obtained the human speech data from YouTube's automatic speech recognition system. Sequences of visual words and speech that are temporally aligned in the original videos are concatenated and fed into a BERT$_\text{base}$ encoder. Similarly, VisualBERT concatenates image region embeddings derived from pretrained object detectors, with their corresponding image captions. The model is pretrained on the COCO~\citep{DBLP:journals/corr/ChenFLVGDZ15} dataset which contains images and five human annotated captions per image. Both pretraining methods rely on the BERT pre-training objectives, modified to their multimodal setups. Specifically, the objectives contain two parts: (1) a masked language modeling (MLM) objective to predict masked out tokens (VideoBERT predicts both visual and text tokens, while VisualBERT predicts only text tokens) and (2) a visual-language prediction objective, which predicts whether the visual and language sequences come from the same video/image or not.   

\paragraph{VL pretraining setup.} For VideoBERT, we obtained the training data and pretrained checkpoints from the authors. For VisualBERT, we downloaded the pretrained VisualBERT-NLVR checkpoint~\footnote{\url{ https://github.com/uclanlp/visualbert}.} pretrained on the Karpathy train split \citep{karpathy2015deep} of COCO \citep{DBLP:journals/corr/ChenFLVGDZ15}. We refer to these two multimodal pretrained checkpoints as VideoBERT$_\text{VL}$ and VisualBERT$_\text{VL}$. Both VideoBERT$_\text{VL}$ and VisualBERT$_\text{VL}$ are based on the BERT$_\text{base}$ architecture, with the difference that our obtained VideoBERT$_\text{VL}$ was trained from scratch (to ensure a controlled comparison to the text-only model, see below), while the public VisualBERT$_\text{VL}$ is initialized with its text-only counterpart.

\paragraph{Text-only pretraining setup.} For comparison, we train text-only counterpart for each model, VideoBERT$_\text{text}$ and VisualBERT$_\text{text}$, using the same text data as the VL model (i.e., the transcribed speech, the captions), while the image data is removed (i.e. the ``visual tokens" of a video or an image). Text-only models are pretrained with masked language modeling objective and next sentence prediction objective, since there are multiple sentences of descriptions of a video (VideoBERT) and multiple captions of an image (VisualBERT). We follow the multimodal pretraining setups as faithfully as possible: we used the same BERT$_\text{base}$ encoder with their corresponding initialization method, the same maximum sequence length, as well as other optimization hyperparameters such as learning rate and number of training epochs. Therefore, the VL models and text-only counterparts have the same architecture and the same number of parameters: VideoBERT models have 125M parameters, while VisualBert models have 109M parameters. More details can be found in Appendix \ref{app:pretraining_details}.

\paragraph{Limitations.} Our experiments are based on two popular variants of VL pretraining frameworks. We picked these two models as they reflect the common trends in VL pretraining for videos and images, and their model architectures and pretraining objectives closely resemble the BERT model, making it easier to compare with their text-only counterparts. However, this comes with the limitation that the models we analyze are trained on data of a different domain than many of our evaluation tasks (e.g., the data for VideoBERT comes from cooking videos on YouTube while the probing tasks are drawn largely from general web text). Thus, absolute results must be interpreted with this domain mismatch in mind. That said, our inclusion of a text-only baseline still allows us to isolate the benefit of the visual modality in an apples-to-apples comparison. Ideally, we would train VL models on multimodal corpora which match the evaluation domains. However, such corpora simply do not exist at the time of writing. Thus, despite the limitations due to domain, our results are representative of the current benefits of VL training. 
\section{Experiments and Results}

We hypothesize that grounded pretraining leads models to learn better linguistic representations than does text-only pretraining. Specifically, we are interested in whether grounded pretraining yields benefits on NLP tasks that are defined entirely over textual inputs, so do not \textit{require} grounded representations (i.e., as opposed to tasks like visual question answering, for which the need for grounded representations is not debatable). 
We consider three different evaluations of ``semantics'': commonsense reasoning about the physical world (\S\ref{sec:piqa}), inferring sentence-level semantic structure (\S\ref{sec:probing}), and composing lexical semantic concepts (\S\ref{sec:clustering}).  

\subsection{Physical Commonsense QA}
\label{sec:piqa}

We first ask whether VL pretraining yields gains to benchmark NLP tasks that intuitively rely on multimodal knowledge, even if they don't explicitly require representing non-text inputs. We use PhysicalQA (PIQA) \citep{bisk2020piqa}, a commonsense reasoning benchmark in which models are given a sentence describing a physical goal (\textit{``Remove gloss from furniture."}) and must select between two candidate solutions (\textit{``Rub furniture with steel wool."}/ \textit{``Rub furniture with cotton ball."}). Following the setup in PIQA, we consider each solution candidate independently by combining the goal with one solution (\verb|[CLS]| goal \verb|[SEP]| solution \verb|[SEP]|), and using the \verb|[CLS]| token embedding at the last hidden layer as the representation of the candidate. We train a probing classifier to perform a binary classification task, with the two candidate representations as its inputs. 

We consider linear, MLP, and transformer probing classifiers. For the linear and MLP probes, we freeze the encoder weights and only train the  classifiers. For the transformer probe, we finetune the last transformer encoder layer and a linear layer on top of it. See Appendix \ref{app:piqa} for details.

\begin{figure}
    \centering
    {%
        \includegraphics[width=\linewidth]{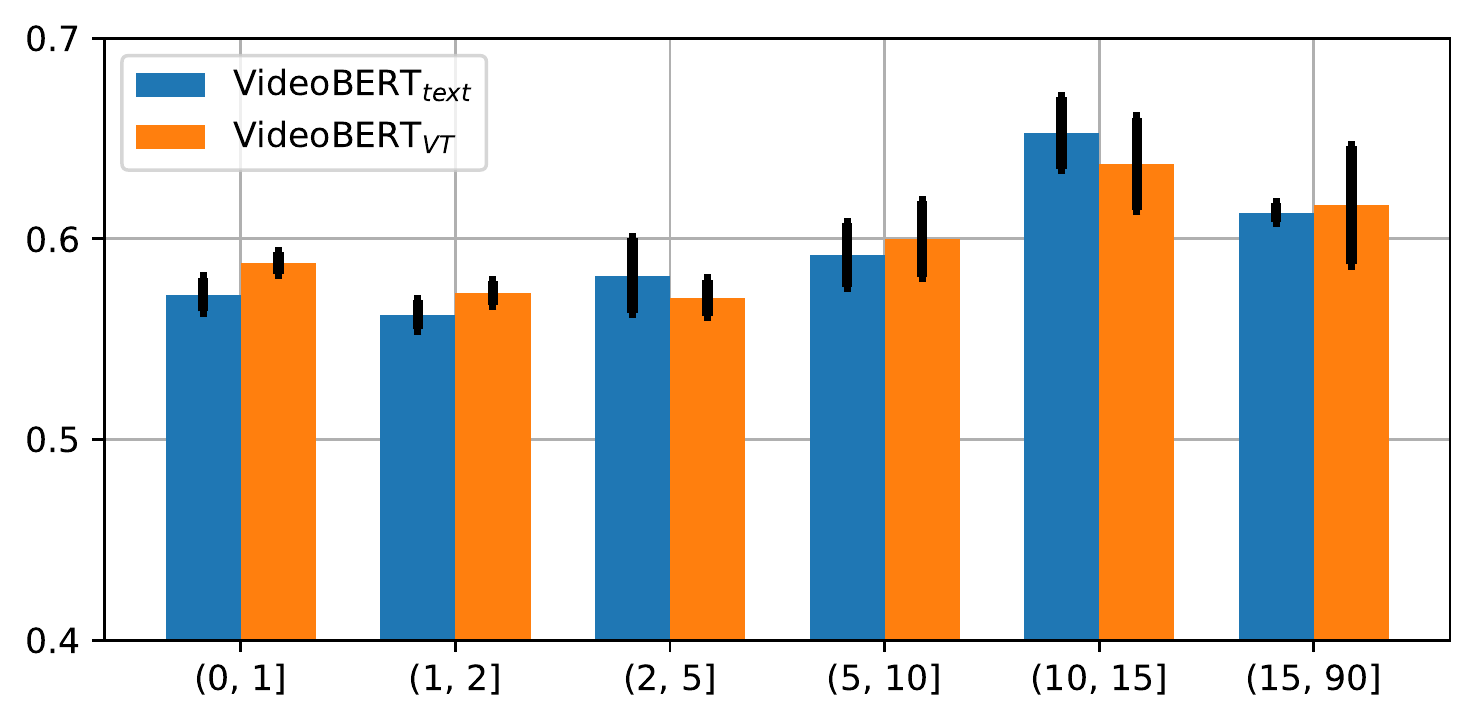}%
        \label{fig:videobert_piqa_bins}%
    }%
    \vfill%
    {%
        \includegraphics[width=\linewidth]{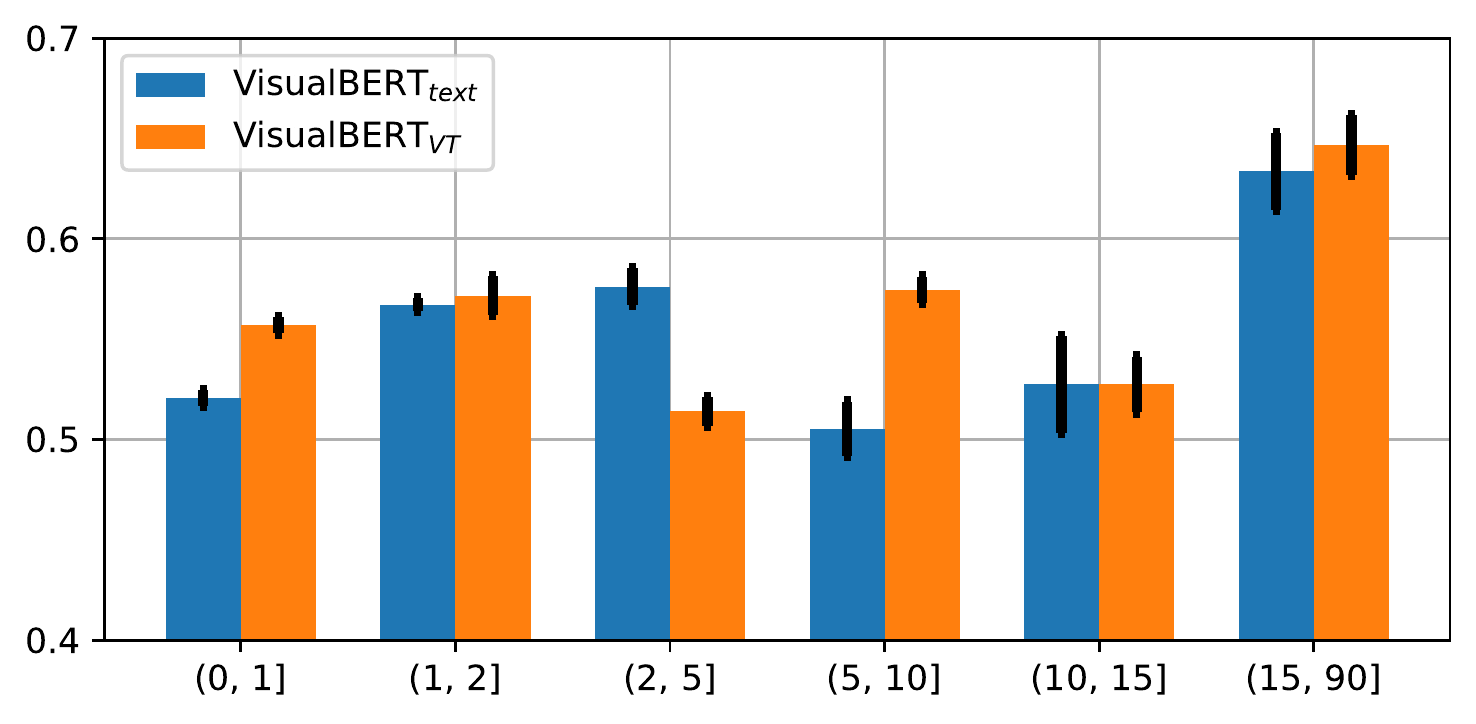}%
        \label{fig:visualbert_piqa_bins}%
    }%
    \caption{\label{fig:piqa_bins} Accuracy divided by word-level edit distance between two solutions. Results are averaged over five runs. Standard deviations are represented by the black lines. VL pretraining brings limited improvements over text-only pretraining on samples with low edit distance.}
\end{figure}

Table \ref{tab:piqa_main} shows our results. Across all settings, we see that VL pretraining produces consistent but marginal gains. In addition, we see that training on YouTube video captions, even without using the video information itself (e.g., comparing VideoBERT$_\text{text}$ to original BERT) yields a few-point improvement. Figure \ref{fig:piqa_bins} shows the results based on word-level edit distance between two solutions. We see that VL pretraining brings a few points improvements when edit distance is low (one or two words), i.e., where picking the right solution hinges on grounded information for single lexical items. On manual inspection of the errors, we do not observe any consistent patterns that reflect different behaviors for VL models and text models. This is true even when we focus only on cooking-related examples for VideoBERT models (i.e., examples we expect to be in domain and thus most likely to demonstrate gains).

Thus, overall, our results are mixed. We see that VL pretraining can yield improvements on text-only tasks, and that these gains likely come from both the difference in the distribution of language as well as the non-linguistic information itself. However, the gains are quite small--only a few points, despite the fact that the task in question (PIQA) is intended to directly probe the type of understanding that one gains from interacting with the physical world. We note, however, that most of the goals and solutions in PIQA are not cooking-related, and thus the limited impact might be due to domain mismatch. Future work on domain-general VL pretraining would offer valuable insight.

\begin{table}[ht!]
    \centering
        \begin{adjustbox}{width=\linewidth, center}
    \begin{tabular}{llrrr}
    \toprule
    \textbf{Encoder}        &   \textbf{Linear}    &   \textbf{MLP}   &   \textbf{Trans.} \\
    \midrule
    BERT$_{\text{base}}$    &   55.43 $\pm$ 0.31   &   57.98 $\pm$ 0.16   &   60.12 $\pm$ 1.43   \\
    \midrule
    VideoBERT$_{\text{text}}$      &   57.87 $\pm$ 0.64   &   58.97 $\pm$ 0.44   &   62.35 $\pm$ 1.23   \\
    VideoBERT$_{\text{VL}}$  &   58.51 $\pm$ 0.20   &   58.56 $\pm$ 0.27   &   63.66 $\pm$ 1.31   \\
    \midrule
    VisualBERT$_{\text{text}}$    &   54.81 $\pm$ 0.19   &   56.81 $\pm$ 0.24   &   58.63 $\pm$ 0.79   \\
    VisualBERT$_{\text{VL}}$  &   55.83 $\pm$ 0.27   &   59.10 $\pm$ 0.11   &   61.66 $\pm$ 1.08   \\
    \bottomrule
    \end{tabular}
    \end{adjustbox}
    \caption{Accuracy $\pm$ std. of different pretrained representations on the validation split of PIQA. Numbers are averaged over five runs. VL pretraining only brings marginal improvements over text-only pretraining.}
    \label{tab:piqa_main}
\end{table}

\subsection{Coreference and Semantic Roles}
\label{sec:probing}

The PIQA results above suggest VL pretraining yields some gains on extrinsic tasks like QA. Such findings invite questions of whether the gains are driven by intrinsic improvements in the semantic representations themselves.
Semantics, after all, is about building intermediate representations that enable the surface form of natural language to ground to entities and events in the real world. 
Thus, intuitively, one might expect that training with explicit access to entities and events would improve such representations (e.g., of predicate-argument structure or entity coreference). 

To test this intuition, we use the edge probing framework from \citet{tenney2018what}, in which a probing classifier takes as input a token span(s), represented as a weighted sum of the layer activations of the token embeddings in the words, and needs to predict a task-related label (e.g. part of speech, parse information). The evaluation suite includes ten syntactic and semantic tasks. Results for all tasks, along with training details, are given in Appendix \ref{app:edge_probing}. Per the above intuition, we are particularly interested in tasks that probe semantic structure. We focus on the following: Entity Coreference (Coref.), e.g., recognizing that \textit{``}\textit{apples}'' and \textit{``them''} refer to the same entity in \textit{``After the apples are chopped, put them in the bowl''}; Semantic Role Labeling (SRL), which requires encoding semantic agents and patients, e.g., recognizing that \textit{``carrots''} are the recipient of the pureeing action in \textit{``The carrots are then pureed in the food processor''}; Semantic Proto-Roles (SPR), which requires predicting features such as \textit{awareness} or \textit{cause} for words in context, e.g., recognizing \textit{``the food processor''} causes the pureeing event, but is not aware of it; and Semantic Relations (Rel.), which requires predicting relations like entity-destination, e.g., the relation between \textit{``apples''} and \textit{``bowl''} in \textit{``put the apples in the bowl''}. 

Table \ref{tab:edge_probing_main} shows results. Across the board, we observe extremely marginal gains in performance when comparing VL models to their text-only counterparts. In $7$ out of $8$ comparisons, the VL model outperforms the text model, versus just $1$ comparison in which the text model outperforms. However, the differences that exist do not appear meaningful ($\sim0.5$ percentage points), and we thus do not conclude that VL pretraining leads to any clear improvement in the models' ability to encode abstract semantic structure. 

\begin{table}[ht!]
    \centering
    \begin{adjustbox}{width=\linewidth, center}
    \begin{tabular}{llrrrr}
    \toprule
    \textbf{Encoder}     &    \textbf{SRL}        &   \textbf{Coref.}  &   \textbf{SPR}  &   \textbf{Rel.}     \\
    \midrule
    BERT$_{\text{base}}$ &   90.10 $\pm$ 0.20   &   95.90 $\pm$ 0.00   &   83.70 $\pm$ 0.00   &   76.25 $\pm$ 0.05   \\
    \midrule
    VideoBERT$_{\text{text}}$  &   84.33 $\pm$ 0.05   &   92.47 $\pm$ 0.05   &   78.23 $\pm$ 0.05   &    65.83 $\pm$ 0.21    \\
    VideoBERT$_{\text{VL}}$  &   84.73 $\pm$ 0.05   &   92.82 $\pm$ 0.05   &   78.80 $\pm$ 0.00   &     66.37 $\pm$ 0.80  \\
    \midrule
    VisualBERT$_{\text{text}}$  & 89.00 $\pm$ 0.00   &   94.87 $\pm$ 0.05   &   82.27 $\pm$ 0.05   &     74.37 $\pm$ 0.19  \\
    VisualBERT$_{\text{VL}}$  &   89.57 $\pm$ 0.21   &   95.13 $\pm$ 0.05   &   82.17 $\pm$ 0.09   &    74.83 $\pm$ 0.05     \\
    \bottomrule
    \end{tabular}
    \end{adjustbox}
    \caption{Comparison of models encoding of various aspects of sentence-level semantic structure (average of micro-averaged F1 score of three runs). We see no significant improvements from VL pretraining. Decreases in performance relative to BERT$_\text{base}$ are likely due to the domain mismatch between image/video captions and the probing evaluation sets (newswire/web text).
    }
    \label{tab:edge_probing_main}
\end{table}

\begin{figure}[ht!]
    \centering
    \begin{subfigure}[b]{\linewidth}
        \includegraphics[width=\linewidth]{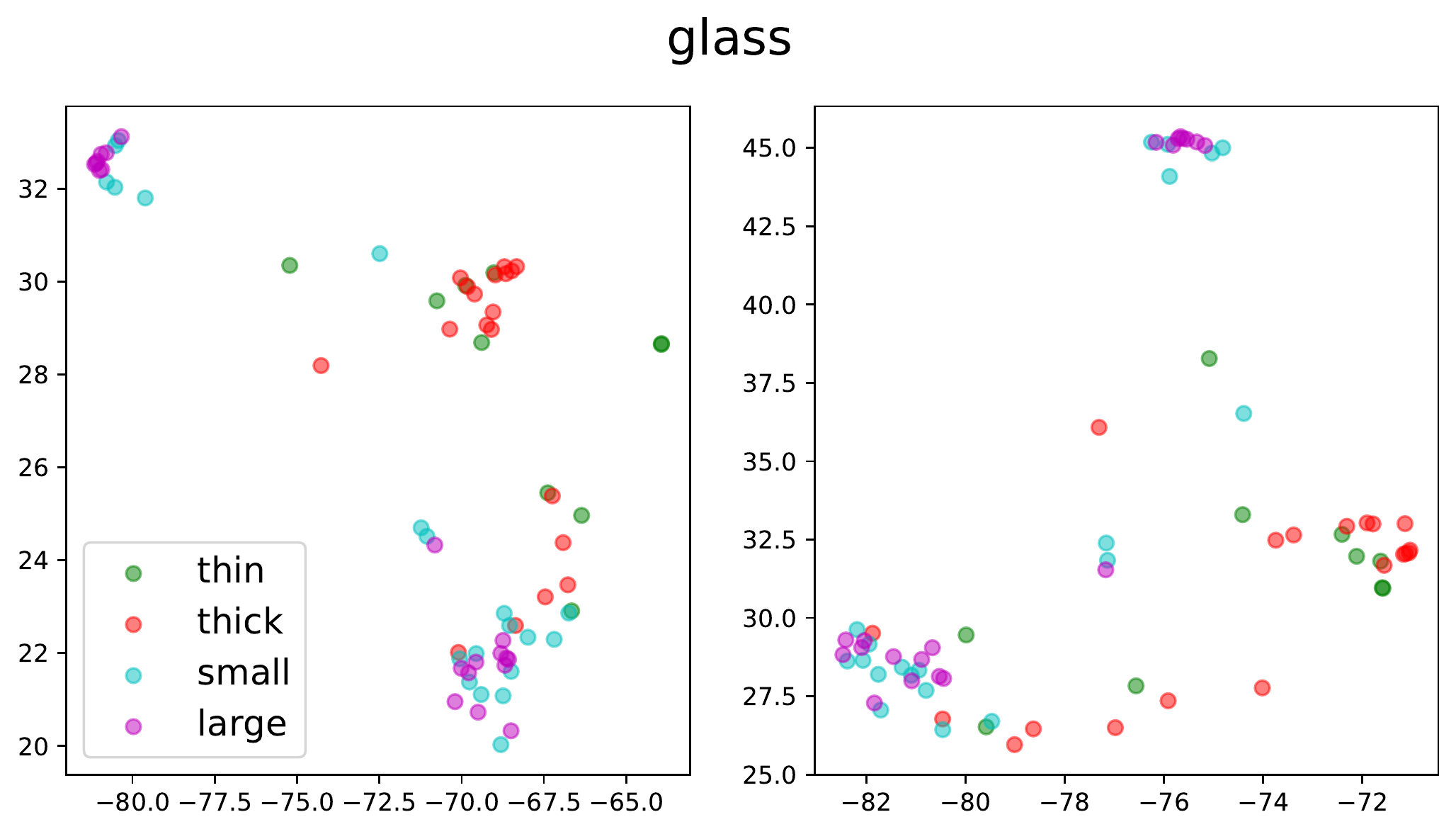}
        \label{fig:clustering_comparison_glass}%
    \end{subfigure}
    \vspace{-2em}
    \caption{TSNE projections of noun representations clusters for ``glass" of VideoBERT$_{\text{text}}$ (Left) and VideoBERT$_{\text{VL}}$ (Right). Neither model distinguishes \textit{small / large} or \textit{thin / thick}.}
    \label{fig:clustering_noun_qualitative}
\end{figure}

\begin{table}[ht!]
    \centering
        \begin{adjustbox}{width=\linewidth, center}
    \begin{tabular}{lrrr}
    \toprule
    \textbf{Encoder}    &  \textbf{Homo.}    &   \textbf{Compl.}   &   \textbf{V-Meas.} \\
    \midrule
    BERT$_{\text{base}}$   &   0.500 $\pm$ 0.017   &   0.526 $\pm$ 0.020   &   0.513 $\pm$ 0.018   \\
    \midrule
    VideoBERT$_{\text{text}}$  &  0.684 $\pm$ 0.025   &   0.702 $\pm$ 0.025   &   0.693 $\pm$ 0.025   \\
    VideoBERT$_{\text{VL}}$   & 0.663 $\pm$ 0.016   &   0.678 $\pm$ 0.016   &   0.670 $\pm$ 0.016   \\
    \midrule
    VisualBERT$_{\text{text}}$  &   0.528 $\pm$ 0.025   &   0.546 $\pm$ 0.028   &   0.537 $\pm$ 0.027   \\
    VisualBERT$_{\text{VL}}$    &   0.546 $\pm$ 0.021   &   0.571 $\pm$ 0.022   &   0.558 $\pm$ 0.022   \\
    \bottomrule
    \end{tabular}
    \end{adjustbox}
    \caption{Summary metrics (range $0$ to $1$) for clustering noun embeddings (e.g., ``apple'') according to their adjective modifiers (e.g., ``ripe''). Numbers are averaged over five random seeds. We see no significant improvement in any metric when grounded (video or image) data is included during training. Homogeneity of 1 means that every point in a cluster belongs to the same class. Completeness of 1 means that every point belonging to a given class is in the same cluster. V-measure is the harmonic mean of the two.}
    \label{tab:clustering_main}
\end{table}

\begin{table}[ht!]
    \centering
    \begin{adjustbox}{width=0.65\linewidth, center}
        \begin{tabular}{llr}
        \toprule
        \textbf{Encoder}    &  \textbf{Accuracy}     \\
        \midrule
        BERT$_{\text{base}}$   &   0.968 $\pm$ 0.002      \\
        \midrule
        VideoBERT$_{\text{text}}$  &  0.992 $\pm$ 0.001      \\
        VideoBERT$_{\text{VL}}$   & 0.993 $\pm$ 0.001      \\
        \midrule
        VisualBERT$_{\text{text}}$  &   0.984 $\pm$ 0.002     \\
        VisualBERT$_{\text{VL}}$    &   0.982 $\pm$ 0.001      \\
        \bottomrule
        \end{tabular}
    \end{adjustbox}
    \caption{Results of probing the noun embeddings to classify the adjectives that modify the nouns. Numbers are averaged over five runs.}
    \label{tab:clustering_adjective_probing}
\end{table}

\subsection{Adjective-Noun Composition}
\label{sec:clustering}

Finally, we investigate whether multimodal pretraining impacts conceptual structure at the lexical level. Arguably, if VL pretraining were to affect linguisitic representations in any meaningful way, we would expect it to manifest in the conceptual representations of visually-groundable concepts. To explore this, we focus on adjective-noun composition, as this provides a simple way of defining a space of visually-groundable objects and properties that we expect conceptual representations to encode. For example, we expect that embeddings of the word {``knife''} from contexts in which the knife is described as {``sharp''} should be more similar to other instances of sharp knives than to instances of knives that are described as {``dull''}. 

We focus on the list of visually grounded adjectives introduced in \citet{isola2015discovering} (e.g., ``small'', ``bright'', ``sharp''). We then mine the WikiHow dataset \citep{DBLP:journals/corr/abs-1810-09305} for all adjective-noun bigrams involving these adjectives. We chose WikiHow because it does not overlap with the training corpus of either of our models, but contains similarly concrete, descriptive language. We perform several additional filters to remove low frequency bigrams, described in Appendix \ref{app:lexical_composition}, which results in an analysis set of 651 unique adjective-noun bigrams across 11,970 contexts. We test how well each pretrained model's representations of the noun (e.g., ``knife'') encodes information about the adjective (e.g., ``sharp'') that modified it. 

Figure \ref{fig:clustering_noun_qualitative} provides a qualitative example of how noun representations cluster when using representations from VideoBERT$_{\text{text}}$ vs.\ VideoBERT$_{\text{VL}}$. Quantitatively (Tables \ref{tab:clustering_main} and \ref{tab:clustering_adjective_probing}; see Appendix \ref{app:lexical_composition} for experimental details), we do not see significant differences between VL and text-only models. Thus, again,VL pretraining does not appear to produce the desired improvements.

\section{Conclusion}

We provide a series of experiments which compare grounded vision-and-language (VL) pretraining to comparable text-only pretraining in terms of the quality of the linguistic representations produced. We find that VL pretraining sometimes produces gains, but that the text-only baselines perform well, and thus the margins are too small to support conclusions that VL pretraining (in its current form) has benefits for NLP in general. While there are good arguments to be made that grounding is necessary for learning general-purpose language representations, we conclude that current methods, which use direct extensions of NLP architectures and are often trained on data from narrow domains, have yet to produce such benefits. Future work is required to explore more domain-general VL training, as well as alternative architectures and losses for combining vision and language signals. 

\section*{Acknowledgments} We would like to thank anonymous reviewers for detailed and helpful comments. We also want to thank Liunian (Harold) Li for helpful feedback on VisualBERT pretraining. This work is supported in part by the DARPA GAILA program. 

\bibliography{anthology,custom}

\begin{thebibliography}{36}
\expandafter\ifx\csname natexlab\endcsname\relax\def\natexlab#1{#1}\fi

\bibitem[{Bender and Koller(2020)}]{bender-koller-2020-climbing}
Emily~M. Bender and Alexander Koller. 2020.
\newblock \href {https://doi.org/10.18653/v1/2020.acl-main.463} {Climbing
  towards {NLU}: {On} meaning, form, and understanding in the age of data}.
\newblock In \emph{Proceedings of the 58th Annual Meeting of the Association
  for Computational Linguistics}, pages 5185--5198, Online. Association for
  Computational Linguistics.

\bibitem[{Bisk et~al.(2020{\natexlab{a}})Bisk, Holtzman, Thomason, Andreas,
  Bengio, Chai, Lapata, Lazaridou, May, Nisnevich, Pinto, and
  Turian}]{bisk-etal-2020-experience}
Yonatan Bisk, Ari Holtzman, Jesse Thomason, Jacob Andreas, Yoshua Bengio, Joyce
  Chai, Mirella Lapata, Angeliki Lazaridou, Jonathan May, Aleksandr Nisnevich,
  Nicolas Pinto, and Joseph Turian. 2020{\natexlab{a}}.
\newblock \href {https://doi.org/10.18653/v1/2020.emnlp-main.703} {Experience
  grounds language}.
\newblock In \emph{Proceedings of the 2020 Conference on Empirical Methods in
  Natural Language Processing (EMNLP)}, pages 8718--8735, Online. Association
  for Computational Linguistics.

\bibitem[{Bisk et~al.(2020{\natexlab{b}})Bisk, Zellers, Gao, Choi
  et~al.}]{bisk2020piqa}
Yonatan Bisk, Rowan Zellers, Jianfeng Gao, Yejin Choi, et~al.
  2020{\natexlab{b}}.
\newblock {PIQA}: Reasoning about physical commonsense in natural language.
\newblock In \emph{Proceedings of the AAAI Conference on Artificial
  Intelligence}, volume~34, pages 7432--7439.

\bibitem[{Brown et~al.(2020)Brown, Mann, Ryder, Subbiah, Kaplan, Dhariwal,
  Neelakantan, Shyam, Sastry, Askell, Agarwal, Herbert{-}Voss, Krueger,
  Henighan, Child, Ramesh, Ziegler, Wu, Winter, Hesse, Chen, Sigler, Litwin,
  Gray, Chess, Clark, Berner, McCandlish, Radford, Sutskever, and
  Amodei}]{DBLP:journals/corr/abs-2005-14165}
Tom~B. Brown, Benjamin Mann, Nick Ryder, Melanie Subbiah, Jared Kaplan,
  Prafulla Dhariwal, Arvind Neelakantan, Pranav Shyam, Girish Sastry, Amanda
  Askell, Sandhini Agarwal, Ariel Herbert{-}Voss, Gretchen Krueger, Tom
  Henighan, Rewon Child, Aditya Ramesh, Daniel~M. Ziegler, Jeffrey Wu, Clemens
  Winter, Christopher Hesse, Mark Chen, Eric Sigler, Mateusz Litwin, Scott
  Gray, Benjamin Chess, Jack Clark, Christopher Berner, Sam McCandlish, Alec
  Radford, Ilya Sutskever, and Dario Amodei. 2020.
\newblock \href {http://arxiv.org/abs/2005.14165} {Language models are few-shot
  learners}.
\newblock \emph{CoRR}, abs/2005.14165.

\bibitem[{Cao et~al.(2020)Cao, Gan, Cheng, Yu, Chen, and Liu}]{cao2020behind}
Jize Cao, Zhe Gan, Yu~Cheng, Licheng Yu, Yen-Chun Chen, and Jingjing Liu. 2020.
\newblock Behind the scene: Revealing the secrets of pre-trained
  vision-and-language models.
\newblock In \emph{European Conference on Computer Vision}, pages 565--580.
  Springer.

\bibitem[{Chen et~al.(2015)Chen, Fang, Lin, Vedantam, Gupta, Doll{\'{a}}r, and
  Zitnick}]{DBLP:journals/corr/ChenFLVGDZ15}
Xinlei Chen, Hao Fang, Tsung{-}Yi Lin, Ramakrishna Vedantam, Saurabh Gupta,
  Piotr Doll{\'{a}}r, and C.~Lawrence Zitnick. 2015.
\newblock \href {http://arxiv.org/abs/1504.00325} {Microsoft {COCO} captions:
  Data collection and evaluation server}.
\newblock \emph{CoRR}, abs/1504.00325.

\bibitem[{Chronis and Erk(2020)}]{chronis-erk-2020-bishop}
Gabriella Chronis and Katrin Erk. 2020.
\newblock \href {https://doi.org/10.18653/v1/2020.conll-1.17} {When is a bishop
  not like a rook? when it{'}s like a rabbi! multi-prototype {BERT} embeddings
  for estimating semantic relationships}.
\newblock In \emph{Proceedings of the 24th Conference on Computational Natural
  Language Learning}, pages 227--244, Online. Association for Computational
  Linguistics.

\bibitem[{Devlin et~al.(2018)Devlin, Chang, Lee, and Toutanova}]{1810.04805}
Jacob Devlin, Ming-Wei Chang, Kenton Lee, and Kristina Toutanova. 2018.
\newblock \href {http://arxiv.org/abs/arXiv:1810.04805} {Bert: Pre-training of
  deep bidirectional transformers for language understanding}.

\bibitem[{Devlin et~al.(2019)Devlin, Chang, Lee, and
  Toutanova}]{devlin-etal-2019-bert}
Jacob Devlin, Ming-Wei Chang, Kenton Lee, and Kristina Toutanova. 2019.
\newblock \href {https://doi.org/10.18653/v1/N19-1423} {{BERT}: Pre-training of
  deep bidirectional transformers for language understanding}.
\newblock In \emph{Proceedings of the 2019 Conference of the North {A}merican
  Chapter of the Association for Computational Linguistics: Human Language
  Technologies, Volume 1 (Long and Short Papers)}, pages 4171--4186,
  Minneapolis, Minnesota. Association for Computational Linguistics.

\bibitem[{Ettinger(2020)}]{ettinger2020}
Allyson Ettinger. 2020.
\newblock \href {https://doi.org/10.1162/tacl\_a\_00298} {What {BERT} is not:
  Lessons from a new suite of psycholinguistic diagnostics for language
  models}.
\newblock \emph{Transactions of the Association for Computational Linguistics},
  8:34--48.

\bibitem[{Forbes et~al.(2019)Forbes, Holtzman, and
  Choi}]{DBLP:journals/corr/abs-1908-02899}
Maxwell Forbes, Ari Holtzman, and Yejin Choi. 2019.
\newblock \href {http://arxiv.org/abs/1908.02899} {Do neural language
  representations learn physical commonsense?}
\newblock \emph{CoRR}, abs/1908.02899.

\bibitem[{Hewitt and Manning(2019)}]{hewitt-manning-2019-structural}
John Hewitt and Christopher~D. Manning. 2019.
\newblock \href {https://doi.org/10.18653/v1/N19-1419} {{A} structural probe
  for finding syntax in word representations}.
\newblock In \emph{Proceedings of the 2019 Conference of the North {A}merican
  Chapter of the Association for Computational Linguistics: Human Language
  Technologies, Volume 1 (Long and Short Papers)}, pages 4129--4138,
  Minneapolis, Minnesota. Association for Computational Linguistics.

\bibitem[{Howell et~al.(2005)Howell, Jankowicz, and Becker}]{howell2005model}
Steve~R Howell, Damian Jankowicz, and Suzanna Becker. 2005.
\newblock A model of grounded language acquisition: Sensorimotor features
  improve lexical and grammatical learning.
\newblock \emph{Journal of Memory and Language}, 53(2):258--276.

\bibitem[{Isola et~al.(2015)Isola, Lim, and Adelson}]{isola2015discovering}
Phillip Isola, Joseph~J Lim, and Edward~H Adelson. 2015.
\newblock Discovering states and transformations in image collections.
\newblock In \emph{Proceedings of the IEEE conference on computer vision and
  pattern recognition}, pages 1383--1391.

\bibitem[{Joshi et~al.(2019)Joshi, Chen, Liu, Weld, Zettlemoyer, and
  Levy}]{joshi2019spanbert}
Mandar Joshi, Danqi Chen, Yinhan Liu, Daniel~S. Weld, Luke Zettlemoyer, and
  Omer Levy. 2019.
\newblock {SpanBERT}: Improving pre-training by representing and predicting
  spans.
\newblock \emph{arXiv preprint arXiv:1907.10529}.

\bibitem[{Karpathy and Fei-Fei(2015)}]{karpathy2015deep}
Andrej Karpathy and Li~Fei-Fei. 2015.
\newblock Deep visual-semantic alignments for generating image descriptions.
\newblock In \emph{Proceedings of the IEEE conference on computer vision and
  pattern recognition}, pages 3128--3137.

\bibitem[{Kingma and Ba(2014)}]{kingma2014adam}
Diederik~P Kingma and Jimmy Ba. 2014.
\newblock Adam: A method for stochastic optimization.
\newblock \emph{arXiv preprint arXiv:1412.6980}.

\bibitem[{Koupaee and Wang(2018)}]{DBLP:journals/corr/abs-1810-09305}
Mahnaz Koupaee and William~Yang Wang. 2018.
\newblock \href {http://arxiv.org/abs/1810.09305} {Wikihow: {A} large scale
  text summarization dataset}.
\newblock \emph{CoRR}, abs/1810.09305.

\bibitem[{Lazaridou et~al.(2015)Lazaridou, Pham, and
  Baroni}]{lazaridou-etal-2015-combining}
Angeliki Lazaridou, Nghia~The Pham, and Marco Baroni. 2015.
\newblock \href {https://doi.org/10.3115/v1/N15-1016} {Combining language and
  vision with a multimodal skip-gram model}.
\newblock In \emph{Proceedings of the 2015 Conference of the North {A}merican
  Chapter of the Association for Computational Linguistics: Human Language
  Technologies}, pages 153--163, Denver, Colorado. Association for
  Computational Linguistics.

\bibitem[{Levine et~al.(2020)Levine, Lenz, Lieber, Abend, Leyton-Brown,
  Tennenholtz, and Shoham}]{levine2020pmimasking}
Yoav Levine, Barak Lenz, Opher Lieber, Omri Abend, Kevin Leyton-Brown, Moshe
  Tennenholtz, and Yoav Shoham. 2020.
\newblock \href {http://arxiv.org/abs/2010.01825} {Pmi-masking: Principled
  masking of correlated spans}.

\bibitem[{Li et~al.(2020)Li, Yatskar, Yin, Hsieh, and
  Chang}]{li-etal-2020-bert-vision}
Liunian~Harold Li, Mark Yatskar, Da~Yin, Cho-Jui Hsieh, and Kai-Wei Chang.
  2020.
\newblock \href {https://doi.org/10.18653/v1/2020.acl-main.469} {What does
  {BERT} with vision look at?}
\newblock In \emph{Proceedings of the 58th Annual Meeting of the Association
  for Computational Linguistics}, pages 5265--5275, Online. Association for
  Computational Linguistics.

\bibitem[{Linzen et~al.(2016)Linzen, Dupoux, and
  Goldberg}]{linzen2016assessing}
Tal Linzen, Emmanuel Dupoux, and Yoav Goldberg. 2016.
\newblock Assessing the ability of {LSTM}s to learn syntax-sensitive
  dependencies.
\newblock \emph{Transactions of the Association for Computational Linguistics},
  4:521--535.

\bibitem[{Majumdar et~al.(2020)Majumdar, Shrivastava, Lee, Anderson, Parikh,
  and Batra}]{vln_bert}
Arjun Majumdar, Ayush Shrivastava, Stefan Lee, Peter Anderson, Devi Parikh, and
  Dhruv Batra. 2020.
\newblock Improving vision-and-language navigation with image-text pairs from
  the web.
\newblock In \emph{European Conference on Computer Vision}, pages 259--274.
  Springer.

\bibitem[{Merrill et~al.(2021)Merrill, Goldberg, Schwartz, and
  Smith}]{DBLP:journals/corr/abs-2104-10809}
William Merrill, Yoav Goldberg, Roy Schwartz, and Noah~A. Smith. 2021.
\newblock \href {http://arxiv.org/abs/2104.10809} {Provable limitations of
  acquiring meaning from ungrounded form: What will future language models
  understand?}
\newblock \emph{CoRR}, abs/2104.10809.

\bibitem[{Radford et~al.(2021)Radford, Kim, Hallacy, Ramesh, Goh, Agarwal,
  Sastry, Askell, Mishkin, Clark, Krueger, and
  Sutskever}]{Radford2021LearningTV}
Alec Radford, Jong~Wook Kim, Chris Hallacy, A.~Ramesh, Gabriel Goh, Sandhini
  Agarwal, Girish Sastry, Amanda Askell, Pamela Mishkin, Jack Clark, Gretchen
  Krueger, and Ilya Sutskever. 2021.
\newblock Learning transferable visual models from natural language
  supervision.
\newblock In \emph{ICML}.

\bibitem[{Radford et~al.(2019)Radford, Wu, Child, Luan, Amodei, and
  Sutskever}]{radford2019language}
Alec Radford, Jeffrey Wu, Rewon Child, David Luan, Dario Amodei, and Ilya
  Sutskever. 2019.
\newblock Language models are unsupervised multitask learners.
\newblock \emph{OpenAI blog}, 1(8):9.

\bibitem[{Rosenberg and Hirschberg(2007)}]{rosenberg2007v}
Andrew Rosenberg and Julia Hirschberg. 2007.
\newblock V-measure: A conditional entropy-based external cluster evaluation
  measure.
\newblock In \emph{Proceedings of the 2007 joint conference on empirical
  methods in natural language processing and computational natural language
  learning (EMNLP-CoNLL)}, pages 410--420.

\bibitem[{Sun et~al.(2019)Sun, Myers, Vondrick, Murphy, and
  Schmid}]{sun2019videobert}
Chen Sun, Austin Myers, Carl Vondrick, Kevin Murphy, and Cordelia Schmid. 2019.
\newblock \href {http://arxiv.org/abs/1904.01766} {Video{BERT}: A joint model
  for video and language representation learning}.

\bibitem[{Tan and Bansal(2020)}]{tan-bansal-2020-vokenization}
Hao Tan and Mohit Bansal. 2020.
\newblock \href {https://doi.org/10.18653/v1/2020.emnlp-main.162}
  {Vokenization: Improving language understanding with contextualized,
  visual-grounded supervision}.
\newblock In \emph{Proceedings of the 2020 Conference on Empirical Methods in
  Natural Language Processing (EMNLP)}, pages 2066--2080, Online. Association
  for Computational Linguistics.

\bibitem[{Tenney et~al.(2019{\natexlab{a}})Tenney, Das, and
  Pavlick}]{tenney-etal-2019-bert}
Ian Tenney, Dipanjan Das, and Ellie Pavlick. 2019{\natexlab{a}}.
\newblock \href {https://doi.org/10.18653/v1/P19-1452} {{BERT} rediscovers the
  classical {NLP} pipeline}.
\newblock In \emph{Proceedings of the 57th Annual Meeting of the Association
  for Computational Linguistics}, pages 4593--4601, Florence, Italy.
  Association for Computational Linguistics.

\bibitem[{Tenney et~al.(2019{\natexlab{b}})Tenney, Xia, Chen, Wang, Poliak,
  McCoy, Kim, Durme, Bowman, Das, and Pavlick}]{tenney2018what}
Ian Tenney, Patrick Xia, Berlin Chen, Alex Wang, Adam Poliak, R~Thomas McCoy,
  Najoung Kim, Benjamin~Van Durme, Sam Bowman, Dipanjan Das, and Ellie Pavlick.
  2019{\natexlab{b}}.
\newblock \href {https://openreview.net/forum?id=SJzSgnRcKX} {What do you learn
  from context? probing for sentence structure in contextualized word
  representations}.
\newblock In \emph{International Conference on Learning Representations}.

\bibitem[{Tenney et~al.(2019{\natexlab{c}})Tenney, Xia, Chen, Wang, Poliak,
  McCoy, Kim, Durme, Bowman, Das, and Pavlick}]{47786}
Ian Tenney, Patrick Xia, Berlin Chen, Alex Wang, Adam Poliak, R.~Thomas McCoy,
  Najoung Kim, Benjamin~Van Durme, Samuel~R. Bowman, Dipanjan Das, and Ellie
  Pavlick. 2019{\natexlab{c}}.
\newblock \href {https://openreview.net/forum?id=SJzSgnRcKX} {What do you learn
  from context? {P}robing for sentence structure in contextualized word
  representations}.
\newblock In \emph{International Conference on Learning Representations}.

\bibitem[{Vuli{\'c} et~al.(2020)Vuli{\'c}, Ponti, Litschko, Glava{\v{s}}, and
  Korhonen}]{vulic-etal-2020-probing}
Ivan Vuli{\'c}, Edoardo~Maria Ponti, Robert Litschko, Goran Glava{\v{s}}, and
  Anna Korhonen. 2020.
\newblock \href {https://doi.org/10.18653/v1/2020.emnlp-main.586} {Probing
  pretrained language models for lexical semantics}.
\newblock In \emph{Proceedings of the 2020 Conference on Empirical Methods in
  Natural Language Processing (EMNLP)}, pages 7222--7240, Online. Association
  for Computational Linguistics.

\bibitem[{Yu and Ettinger(2020)}]{yu-ettinger-2020-assessing}
Lang Yu and Allyson Ettinger. 2020.
\newblock \href {https://doi.org/10.18653/v1/2020.emnlp-main.397} {Assessing
  phrasal representation and composition in transformers}.
\newblock In \emph{Proceedings of the 2020 Conference on Empirical Methods in
  Natural Language Processing (EMNLP)}, pages 4896--4907, Online. Association
  for Computational Linguistics.

\bibitem[{Zellers et~al.(2018)Zellers, Bisk, Schwartz, and
  Choi}]{zellers-etal-2018-swag}
Rowan Zellers, Yonatan Bisk, Roy Schwartz, and Yejin Choi. 2018.
\newblock \href {https://doi.org/10.18653/v1/D18-1009} {{SWAG}: A large-scale
  adversarial dataset for grounded commonsense inference}.
\newblock In \emph{Proceedings of the 2018 Conference on Empirical Methods in
  Natural Language Processing}, pages 93--104, Brussels, Belgium. Association
  for Computational Linguistics.

\bibitem[{Zellers et~al.(2019)Zellers, Holtzman, Bisk, Farhadi, and
  Choi}]{zellers-etal-2019-hellaswag}
Rowan Zellers, Ari Holtzman, Yonatan Bisk, Ali Farhadi, and Yejin Choi. 2019.
\newblock \href {https://doi.org/10.18653/v1/P19-1472} {{H}ella{S}wag: Can a
  machine really finish your sentence?}
\newblock In \emph{Proceedings of the 57th Annual Meeting of the Association
  for Computational Linguistics}, pages 4791--4800, Florence, Italy.
  Association for Computational Linguistics.

\end{thebibliography}
\bibliographystyle{acl_natbib}

\newpage
\appendix
\section{VL and Text-only Pretraining}

\subsection{Domain-specific Masking}
Masking tokens uniformly at random in BERT is found to be suboptimal \citep{joshi2019spanbert, levine2020pmimasking}. In addition, we hypothesize that the benefits of visual-linguistic alignment might be greater if masking occurs on content words (which, in the cooking domain, are likely to be visually-groundable concepts). Thus, we implement a \textit{domain-specific masking}, which aggressively masks the most frequent cooking-related verbs and nouns. We apply the BERT tokenizer on the cooking corpus, and manually pick the most frequent 500 cooking-related tokens. During the pretraining data generation, 15\% of the tokens will be chosen, where the frequent tokens has 80\% probability of being chosen, while the other tokens has 15\% probability. The masking strategy is similar to the original BERT, where 80\%/10\%/10\% of the chosen tokens will be replaced with \verb|[MASK]| tokens/ random tokens/ the original tokens respectively. VideoBERT is pretrained with both random masking and domain-specific masking, while VisualBERT is only pretrained with random masking.

\subsection{Pretraining Details} \label{app:pretraining_details}
We pretrain VideoBERT$_\text{text}$ from scratch on the same cooking dataset in \citep{sun2019videobert}. We strictly follow the training setup of VideoBERT$_\text{VL}$ which is based on BERT$_\text{base}$: it has 12 layers of transformer blocks, where each block has 768 hidden units and 12 self-attention heads. We use 4 Cloud TPUs with a total batch size of 128, and we train a model for 400K iterations. We use the Adam optimizer with an initial learning rate of 1e-5, and a linear decay learning rate schedule. The training process takes around 2 days.

We initialize VisualBERT$_\text{text}$ with the pretrained BERT$_\text{base}$ weights released by \citep{1810.04805}. This text-only model has the same configuration as its VL variant: it has 12 layers of transformer blocks, where each block has 768 hidden units and 12 self-attention heads. 
The training process also largely follows the setup of VisualBERT$_\text{VL}$: we use 4 TitanV GPUs with a total batch size of 64 and cap the sequences whose lengths are longer than 128. VisualBERT$_\text{text}$ is trained for 10 epochs, or roughly 90K iterations, with the Adam optimizer with an initial learning rate of 5e-5. The warm-up step number is set to 10\% of the total training step count. The training process takes around 25 hours.
 
\section{Experimental Details}

\subsection{PIQA} \label{app:piqa}

We use \verb|[CLS]| token embedding $e$ at the last hidden layer as the representation of a candidate solution for a goal. This embedding will be passed into the probing classifiers: a single linear layer, an MLP, and a transformer. The MLP probe has a hidden size of 512 and has architecture as below:
\begin{align*}
    h &= \mathtt{tanh}(W_1 e + b_1) \\
    h &= \mathtt{layer\_norm}(h) \\
    \text{output} &= W_2 h + b_2
\end{align*}
We train a model by cross-entropy loss and by using the Adam optimizer \citep{kingma2014adam} with a batch size of 32, an initial learning rate of 1e-4. We evaluate a model on the validation set every 1000 steps, halve the learning rate if no improvement is seen in 5 validations, and stop training if no improvement is seen in 20 validations. In this way, we limit the expressive power of the probes (since we are primarily interested in understanding differences in the representations that result directly from pretraining), yet still consider a number of ways (linear/nonlinear) that such information could potentially be encoded. 

\begin{table*}[ht!]
    \centering
    \begin{adjustbox}{width=\linewidth, center}
    \begin{tabular}{ll|l|ccc|ccc|ccc}
    \toprule
    
    \multirow{2}{*}{\textbf{Encoder}}      &   \multirow{2}{*}{\textbf{Masking}}   &     \textbf{Batch Size}    &   \multicolumn{3}{|c|}{\textbf{32}}    &   \multicolumn{3}{c|}{\textbf{64}}  &  \multicolumn{3}{c}{\textbf{128}}  \\
    \cmidrule{3-12}
    
    &   &     \textbf{LR}    &   \textbf{1e-4}    &   \textbf{5e-5}       &   \textbf{2e-5}    &   \textbf{1e-4}    &   \textbf{5e-5}       &   \textbf{2e-5}   &   \textbf{1e-4}    &   \textbf{5e-5}       &   \textbf{2e-5}  \\
    
    \midrule
    BERT$_{\text{base}}$ & Random   &   & 60.12 $\pm$ 1.43   &   60.61 $\pm$ 1.82   &   58.08 $\pm$ 1.34   &   59.17 $\pm$ 1.37   &   58.14 $\pm$ 1.31   &   57.21 $\pm$ 0.39   &   59.58 $\pm$ 1.57   &   58.18 $\pm$ 0.75   &   56.79 $\pm$ 0.30      \\
    \midrule
    VideoBERT$_\text{text}$  & Random       &  &  62.02 $\pm$ 0.73   &   60.10 $\pm$ 1.07   &   58.45 $\pm$ 1.63   &   61.35 $\pm$ 1.31   &   60.89 $\pm$ 0.58   &   57.94 $\pm$ 0.91   &   61.06 $\pm$ 0.95   &   59.89 $\pm$ 0.50   &   56.40 $\pm$ 0.51   \\
    VideoBERT$_\text{VL}$ & Random &  & 63.66 $\pm$ 1.31   &   62.68 $\pm$ 0.76   &   60.51 $\pm$ 0.66   &   62.77 $\pm$ 0.80   &   63.13 $\pm$ 0.35   &   59.60 $\pm$ 0.96   &   62.59 $\pm$ 0.66   &   61.39 $\pm$ 0.40   &   59.65 $\pm$ 0.73      \\
    \midrule
    VideoBERT$_\text{text}$  & Domain          &   &   61.66 $\pm$ 0.64   &   60.04 $\pm$ 0.57   &   58.97 $\pm$ 0.63   &   60.66 $\pm$ 0.87   &   60.42 $\pm$ 1.13   &   58.43 $\pm$ 1.15   &   60.18 $\pm$ 0.98   &   58.48 $\pm$ 0.53   &   57.62 $\pm$ 1.58      \\
    VideoBERT$_\text{VL}$  & Domain    & &  62.30 $\pm$ 0.95   &   61.10 $\pm$ 0.43   &   59.67 $\pm$ 0.40   &   61.73 $\pm$ 0.53   &   60.67 $\pm$ 0.74   &   59.26 $\pm$ 0.46   &   61.48 $\pm$ 0.60   &   59.99 $\pm$ 1.21   &   58.03 $\pm$ 1.18   \\
    \midrule
    VisualBERT$_\text{text}$  & Random      &      &   58.63 $\pm$ 0.79   &   57.95 $\pm$ 0.71   &   56.09 $\pm$ 1.20   &   58.65 $\pm$ 0.86   &   57.69 $\pm$ 0.56   &   55.32 $\pm$ 0.75   &   58.03 $\pm$ 1.16   &   57.81 $\pm$ 0.83   &   55.51 $\pm$ 0.90      \\
    VisualBERT$_\text{VL}$  & Random    &   &   61.66 $\pm$ 1.08   &   60.78 $\pm$ 0.38   &   59.30 $\pm$ 0.96   &   60.70 $\pm$ 0.55   &   60.55 $\pm$ 0.66   &   59.09 $\pm$ 0.79   &   60.40 $\pm$ 0.72   &   60.07 $\pm$ 0.18   &   58.52 $\pm$ 1.04   \\
    \bottomrule
    \end{tabular}
    \end{adjustbox}
    \caption{Hyperparameter search for transformer probing models on PIQA. LR refers to learning rate. Numbers are average accuracy of five runs. VL checkpoints consistently outperform text-only counterparts by a small margin. This rules out the likelihood that the improvement of VL pretraining is caused by specific hyperparameter settings.}
    \label{tab:piqa_transformer_hparam_search}
\end{table*}

\begin{table*}[ht!]
    \centering
    \begin{adjustbox}{width=\linewidth, center}
    \begin{tabular}{llcccccccccc}
    \toprule
    
    \textbf{Encoder}      &   \textbf{Masking}   &     \textbf{POS}    &   \textbf{Consti.}    &   \textbf{Dep.}       &   \textbf{NER}    &   \textbf{SRL}        &   \textbf{Coref.$^\text{O}$}  &   \textbf{SPR1}   &   \textbf{SPR2}   &   \textbf{Coref.$^\text{W}$}    &   \textbf{SemEval}  \\
    \midrule
    BERT$_{\text{base}}$ & Random   &   96.47 $\pm$ 0.05    &   86.80 $\pm$ 0.14   &   95.20 $\pm$ 0.08   &   96.00 $\pm$ 0.00   &   90.10 $\pm$ 0.20   &   95.90 $\pm$ 0.00   &   83.70 $\pm$ 0.00   &   82.80 $\pm$ 0.08   &   57.90 $\pm$ 0.00   &   76.25 $\pm$ 0.05   \\
    \midrule
    VideoBERT$_\text{text}$  & Random       &   93.90  $\pm$ 0.28   &   82.30 $\pm$ 0.14   &   92.07 $\pm$ 0.05   &   91.60 $\pm$ 0.14   &   84.33 $\pm$ 0.05   &   92.47 $\pm$ 0.05   &   78.23 $\pm$ 0.05   &   81.30 $\pm$ 0.00   &   56.20 $\pm$ 0.14   &   65.83 $\pm$ 0.21   \\
    VideoBERT$_\text{VL}$ & Random &   93.87 $\pm$ 0.09   &   83.50 $\pm$ 0.00   &   92.27 $\pm$ 0.05   &   92.00 $\pm$ 0.14   &   84.73 $\pm$ 0.05   &   92.82 $\pm$ 0.05   &   78.80 $\pm$ 0.00   &   81.30 $\pm$ 0.00   &   56.20 $\pm$ 1.98   &   66.37 $\pm$ 0.80   \\
    \midrule
    VideoBERT$_\text{text}$  & Domain          &   93.10  $\pm$ 0.00   &   82.20 $\pm$ 0.00   &   91.10 $\pm$ 0.16   &   90.77 $\pm$ 0.33   &   82.93 $\pm$ 0.05   &   92.40 $\pm$ 0.00   &   76.83 $\pm$ 0.12   &   81.03 $\pm$ 0.05   &   54.43 $\pm$ 0.90   &   62.97 $\pm$ 0.47   \\
    VideoBERT$_\text{VL}$  & Domain    &   93.33 $\pm$ 0.09   &   82.37 $\pm$ 0.09   &   91.47 $\pm$ 0.09   &   90.67 $\pm$ 0.17    &   82.93 $\pm$ 0.05   &   92.30 $\pm$ 0.00   &   78.03 $\pm$ 0.05   &   81.13 $\pm$ 0.09   &   56.07 $\pm$ 0.52   &   64.63 $\pm$ 0.19   \\
    \midrule
    VisualBERT$_\text{text}$  & Random          &   95.40 $\pm$ 0.00   &   86.20 $\pm$ 0.14   &   94.20 $\pm$ 0.00   &   94.60 $\pm$ 0.57   &   89.00 $\pm$ 0.00   &   94.87 $\pm$ 0.05   &   82.27 $\pm$ 0.05   &   82.40 $\pm$ 0.00   &   57.57 $\pm$ 2.03   &   74.37 $\pm$ 0.19   \\
    VisualBERT$_\text{VL}$  & Random    &   96.10 $\pm$ 0.14   &   86.23 $\pm$ 0.05   &   94.57 $\pm$ 0.05   &   95.20 $\pm$ 0.00   &   89.57 $\pm$ 0.21   &   95.13 $\pm$ 0.05   &   82.17 $\pm$ 0.09   &   82.43 $\pm$ 0.05   &   58.13 $\pm$ 0.61   &   74.83 $\pm$ 0.05   \\
    \bottomrule
    \end{tabular}
    \end{adjustbox}
    \caption{Results of models on a suite of edge probing tasks. Numbers reported are average of micro-averaged F1 score of three runs. We see no significant improvements when the models are pretrained with both text and visual data.}
    \label{tab:edge_probing_full}
\end{table*}

\begin{figure}[ht!]
    \centering
    \includegraphics[width=1\linewidth]{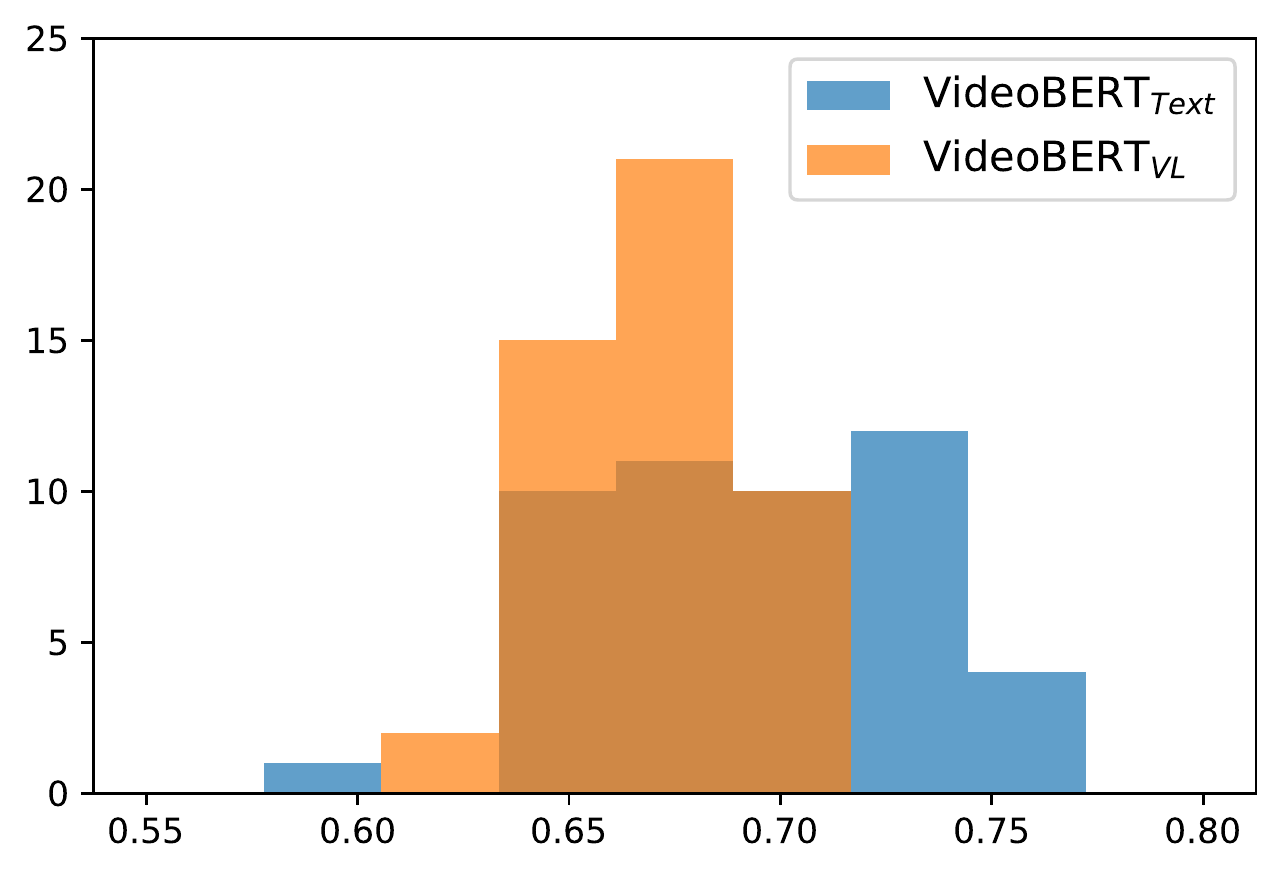}
    \caption{Histograms of V-measure scores for clustering noun embeddings according to their adjective modifiers of VideoBERT$_{\text{text}}$ (Random) and VideoBERT$_{\text{VL}}$ (Random).}
    \label{fig:clustering_noun_distribution_v_measure}
\end{figure}

\subsection{Syntactic and Semantic ``Edge Probing'' Tasks} \label{app:edge_probing}

Edge probing formulates probing tasks into the same format, where the probing classifier takes a span $s_1 = [i_1, j_1)$ and an optional span $s_2 = [i_2, j_2)$, and needs to predict a task-related label based on the span representations. A span representation is a weighted sum of the layer activations of the token embeddings in the given spans. We train a probing classifier for each task with encoder weights frozen, and follow the probing architecture and training strategy in \citep{47786}. Figure~\ref{tab:edge_probing_full} shows the results on all tasks for all models. 

\begin{table}
    \centering
        \begin{adjustbox}{width=\linewidth, center}
    \begin{tabular}{llrrr}
    \toprule
    \textbf{Encoder}      &    \textbf{Masking}    &  \textbf{Homo.}    &   \textbf{Compl.}   &   \textbf{V-Meas.} \\
    \midrule
    BERT$_{\text{base}}$   & Random &   0.500 $\pm$ 0.017   &   0.526 $\pm$ 0.020   &   0.513 $\pm$ 0.018   \\
    \midrule
    VideoBERT$_{\text{text}}$  & Random &  0.684 $\pm$ 0.025   &   0.702 $\pm$ 0.025   &   0.693 $\pm$ 0.025   \\
    VideoBERT$_{\text{VL}}$   & Random & 0.663 $\pm$ 0.016   &   0.678 $\pm$ 0.016   &   0.670 $\pm$ 0.016   \\
    \midrule
    VideoBERT$_{\text{text}}$  & Domain &  0.409 $\pm$ 0.014   &   0.442 $\pm$ 0.013   &   0.425 $\pm$ 0.014   \\
    VideoBERT$_{\text{VL}}$    & Domain  & 0.393 $\pm$ 0.031   &   0.422 $\pm$ 0.031   &   0.407 $\pm$ 0.031   \\
    \midrule
    VisualBERT$_{\text{text}}$  &   Random  &   0.528 $\pm$ 0.025   &   0.546 $\pm$ 0.028   &   0.537 $\pm$ 0.027   \\
    VisualBERT$_{\text{VL}}$    &   Random  &   0.546 $\pm$ 0.021   &   0.571 $\pm$ 0.022   &   0.558 $\pm$ 0.022   \\
    \bottomrule
    \end{tabular}
    \end{adjustbox}
    \caption{Summary metrics (range $0$ to $1$) for clustering noun embeddings (e.g., ``apple'') according to their adjective modifiers (e.g., ``ripe''). Numbers are averaged over five random seeds. We see no significant improvement in any metric when grounded (video or image) data is included during training. Homogeneity of 1 means that every point in a cluster belongs to the same class. Completeness of 1 means that every point belonging to a given class is in the same cluster. V-measure is the harmonic mean of the two.}
    \label{tab:clustering_main_appendix}
\end{table}

\begin{table}
    \centering
    \begin{adjustbox}{width=0.7\linewidth, center}
        \begin{tabular}{llr}
        \toprule
        \textbf{Encoder}      &    \textbf{Masking}    &  \textbf{Accuracy}     \\
        \midrule
        BERT$_{\text{base}}$   & Random &   0.968 $\pm$ 0.002      \\
        \midrule
        VideoBERT$_{\text{text}}$  & Random &  0.992 $\pm$ 0.001      \\
        VideoBERT$_{\text{VL}}$   & Random & 0.993 $\pm$ 0.001      \\
        \midrule
        VideoBERT$_{\text{text}}$  & Domain &  0.964 $\pm$ 0.002     \\
        VideoBERT$_{\text{VL}}$    & Domain  & 0.973 $\pm$ 0.001     \\
        \midrule
        VisualBERT$_{\text{text}}$  &   Random  &   0.984 $\pm$ 0.002     \\
        VisualBERT$_{\text{VL}}$    &   Random  &   0.982 $\pm$ 0.001      \\
        \bottomrule
        \end{tabular}
    \end{adjustbox}
    \caption{Results of probing the noun embeddings to classify the adjectives that modify the nouns. Numbers are averaged over five runs.}
    \label{tab:clustering_adjective_probing_appendix}
\end{table}

\subsection{Lexical Composition} \label{app:lexical_composition}

We preprocess WikiHow dataset by tokenizing the 215K instructions into 5 million single sentences. We run a bigram search over all the sentences to find pairs of an adjective and a noun. The lower bound of bigram occurrence is set to 10, while the bigrams whose nouns do not pair with more than 10 unique adjectives are filtered out. Eventually, this leaves us 57,521 bigrams and 651 unique bigrams. Encoders then produce the representations of the nouns in these bigrams.

Following, we apply a visually grounded adjective filter based on the list of adjectives introduced in \citep{isola2015discovering}. For a unique bigram, up to 20 noun representations are randomly sampled. Finally, there are 62 unique adjectives, 48 unique nouns, and 11,970 noun representations. 

We use K-Means to cluster the representations of each noun, with $K$ equal to the the number of unique adjectives that modifies the noun in our dataset. We measure the quality of the resulting clusters using three clustering metrics: homogeneity, completeness, and V-measure~\footnote{The ratio of weight attributed to homogeneity and completeness is set to 1:1.}\citep{rosenberg2007v}, which are roughly analogous to precision, recall, and f1-score. We use the adjectives as the ground-truth cluster labels; i.e., scores are higher when the noun representations cluster according to the adjectives which modifies the noun in context. 

Last, we carry out a probing experiment to attempt to evaluate the adjective information that is linearly encoded in the noun representations produced by the models. Given noun embeddings, a linear probing classifier, that is built on top of each model, classifies the adjectives that modify the nouns.

Based on a series of quantitative analyses, Tables \ref{tab:clustering_main_appendix} and \ref{tab:clustering_adjective_probing_appendix} and Figure \ref{fig:clustering_noun_distribution_v_measure}, we do not see significant differences between VL and text-only models.

\end{document}